%% file: regular.tex
\documentclass{article}
\usepackage{spconf,amsmath,graphicx,tabularx,blindtext,color,url,algorithm}
\usepackage[noend]{algpseudocode}
\usepackage[hang,flushmargin]{footmisc}

\title{SPEAKER ADAPTIVE TRAINING USING MODEL AGNOSTIC META-LEARNING}
\name{Ond\v{r}ej Klejch, Joachim Fainberg, Peter Bell, Steve Renals\thanks{This work was partially supported by: the EU H2020 projects SUMMA (grant agreement 688139) and ELG (grant agreement 825627), and a PhD studentship funded by Bloomberg.}}
\address{Centre for Speech Technology Research, University of Edinburgh, United Kingdom}

\def\adapt{adapt}

\begin{document}

\maketitle

\begin{abstract}

\input{abstract.tex}

\end{abstract}

\begin{keywords}
speaker adaptation, speaker adaptive training, model-agnostic meta-learning
\end{keywords}

\input{text.tex}

\bibliographystyle{IEEEbib}
\bibliography{mybib}

\end{document}

%% file: abstract.tex
Speaker adaptive training (SAT) of neural network acoustic models learns models in a way that makes them more suitable for adaptation to test conditions. Conventionally, model-based speaker adaptive training is performed by having a set of speaker dependent parameters that are jointly optimised with speaker independent parameters in order to remove speaker variation. However, this does not scale well if  all neural network weights are to be adapted to the speaker. In this paper we formulate speaker adaptive training as a meta-learning task, in which an adaptation process using gradient descent is encoded directly into the training of the model. We compare our approach with test-only adaptation of a standard baseline model and a SAT-LHUC model with a learned speaker adaptation schedule and demonstrate that the meta-learning approach achieves comparable results.

%% file: text.tex
\section{Introduction}

Adapting an acoustic model to unseen test speakers can improve the accuracy of an automatic speech recognition (ASR) system by reducing the mismatch between training and test conditions. Speaker adaptive training (SAT)~\cite{anastasakos1996compact} further reduces the mismatch by removing speaker variance during training of the acoustic model, thus allowing it to focus solely on modelling phonetic variations. Speaker adaptation approaches can be divided into three groups: \textit{Feature-space} methods estimate a  transformation of the acoustic features to improve performance on the test  data~\cite{leggetter1995maximum,gales1998maximum}; \textit{Model-based} methods adapt the weights of an acoustic model~\cite{swietojanski2016learning, xue2014singular, liao2013speaker, yu2013kl,wang2017unsupervised}; and \textit{Hybrid} methods use auxiliary features to inform the acoustic model about speaker identity~\cite{saon2013speaker,abdel2013fast}. These approaches are complementary and can be used together~\cite{samarakoon2016combining}.

Speaker adaptive training may be applied to all these adaptation approaches. In  feature-space adaptation, training of the acoustic model weights is interleaved with estimation of speaker-dependent feature transformations in order to remove speaker variability and to focus the acoustic model on  phonetic variations~\cite{anastasakos1996compact}. In model-based speaker adaptive training,  the acoustic model is parameterised as speaker-dependent and speaker-independent weights. A copy of the speaker-dependent weights is maintained and optimised separately for each speaker during the training process in order to factor out speaker variation from the canonical speaker-independent acoustic model~\cite{ochiai2015speaker,ochiai2016bottleneck,swietojanski2016sat}. Finally, all hybrid approaches can be considered as speaker adaptive training because they provide information about speaker identity, which allows the acoustic model to easily remove speaker variation from the input features~\cite{saon2013speaker,abdel2013fast,miao2015speaker}. In this paper we focus on model-based adaptation and speaker adaptive training for model-based adaptation. 

The biggest challenge of speaker adaptation is to improve performance on test data as much as possible without overfitting to test transcripts. This is especially important in a rapid adaptation setting when we use only a small amount of adaptation data to adapt the weights of an acoustic model. Therefore, model-based speaker adaptation usually adapts only a small number of speaker-dependent weights of an acoustic model in order not to overfit to the adaptation data. Traditionally, activations of hidden layers are adapted using linear transformations. Depending on whether we adapt inputs, hidden activations or outputs we talk about Linear Input Network (LIN), Linear Hidden Network (LHN) or Linear Output Network (LON) respectively~\cite{gemello2006adaptation,gemello2007linear}. Even then these linear transformations might use too many weights, therefore low-rank approximations may be used previously~\cite{xue2014singular,zhao2017extended}. Learning Hidden Unit Contributions (LHUC)~\cite{swietojanski2016learning, swietojanski2014learning} is a special case of such a transformation, in which we force the speaker dependent matrix to be diagonal, thus learning a magnitude for each hidden unit. Another similar approach adapts the scale and offset of the batch normalisation layers~\cite{wang2017unsupervised}.

The methods described above require considerable manual work to carefully select which weights should be treated as speaker dependent. Moreover, these methods are not able to perform speaker adaptive training of all weights of the acoustic model which limits the expressivity of the speaker dependent transformations. This is because it is not possible to store all weights of the acoustic model for each speaker in memory and we usually do not have enough speaker dependent training data to reliably train all those weights. Furthermore, even though the acoustic model was trained in a speaker adaptive way using these methods, we still need to find a reliable adaptation schedule to perform well in speaker adaptation. Although these problems might be solved with a careful hyper-parameter search, we can address them more explicitly by formulating the problem as a \textit{meta-learning} task. 

The aim of meta-learning is to replace handcrafted algorithms with learned algorithms, similar to how deep learning replaced handcrafted features in traditional machine learning. Recently, meta-learning has been successfully used for learning task specific learning schedules that produce better results than traditional learning schedules~\cite{andrychowicz2016learning}, learning good initialisations for few-shot classifiers~\cite{ravi2016optimization,finn2017model} or language transfer in low-resource neural machine translation~\cite{gu2018meta}.

Previously speaker adaptation was formulated as a meta-learning task~\cite{klejch2018learning}. In this paper we extend that approach to perform speaker adaptive training using a method called Model Agnostic Meta-Learning (MAML)~\cite{finn2017model}. This is a new perspective on speaker adaptive training because rather than using speaker dependent weights to remove speaker variation from the data, we are directly encoding the gradient-descent based speaker adaptation process into the training of the acoustic model. Our objective is to find weights more amenable to speaker adaptation by steering the training of the model to find regions that are  better suited to rapid adaptation.

The contributions of this paper are:
\begin{enumerate}
    \item We experimentally show that a baseline state-of-the-art acoustic model adapted with a schedule learned with MAML achieves significantly better results than the handcrafted adaptation schedule used in~\cite{klejch2018learning}. 
    \item We explain how to adapt models that use batch normalisation~\cite{ioffe2015batch}, and show that it is better to use global statistics rather than batch statistics during adaptation.
    \item We show that speaker adaptive training using MAML achieves comparable results with the baseline approaches.
\end{enumerate}



\section{Adaptation as Meta-Learning}

Speaker adaptation has previously been formulated as a meta-learning task~\cite{klejch2018learning}. In this section we review this formulation and show how it can be extended to speaker adaptive training.

\subsection{Speaker Adaptation as a Meta-Learning Task}

The goal of speaker adaptation is to adapt an acoustic model $f(x; \Theta)$ with weights $\Theta$ to perform better on a speaker $a$ using adaptation data $D_a = (X_a, Y_a)$. Speaker adaptation may be formulated as a function that given an acoustic model $f(x; \Theta)$ and adaptation data $D_a$ produces adapted weights $\Theta'$:
\begin{equation}
    adapt(f, \Theta, D_a) \rightarrow \Theta' \, .
\end{equation}
Depending on the scenario, the labels $Y_a$, corresponding to the acoustic input $X_a$, might be obtained from a reference transcript (supervised adaptation) or obtained from the best path of a first pass decode (unsupervised adaptation). 

The performance of an acoustic model on unseen data $D_u = (X_u, Y_u)$ is measured as the value of a loss function:
\begin{equation}
    L(Y_u, f(X_u; \Theta)),
\end{equation}
for example categorical cross-entropy, frame error rate or word error rate (WER). Note that the labels $Y_u$ are always obtained from the reference transcripts since we want to measure the true performance of the model. Similarly, we measure the loss of an adapted acoustic model by:
\begin{equation}
    L(Y_u, f(X_u; \adapt(f, \Theta, D_a)).
    \label{eq:adapted_loss}
\end{equation}

To train the adaptation function using the meta-learning approach, we require the function to be both parametric and differentiable. We therefore add parameters $\Phi$ to the adaptation function (Our choice of $\Phi$ will be defined in Section~2.3.):
\begin{equation}
    adapt(f, \Theta, D_a; \Phi) \rightarrow \Theta' \, .
\end{equation}

We are now ready to introduce the loss of the meta-learner. Recall that the goal of speaker adaptation is to adapt an acoustic model $f(x; \Theta)$ using adaptation data $D_a$ in order to improve performance on unseen data $D_u$. The loss of the meta-learner can then be expressed as a sum of losses of the adapted models~(Equation~\ref{eq:adapted_loss}) over a meta set $\mathcal{D}$:
\begin{equation}
    J = \sum_\mathcal{D} L(Y_u, f(X_u; \adapt(f, \Theta, D_a; \Phi))),
    \label{loss}
\end{equation}
where $\mathcal{D}$ consists of tuples of adaptation data for the adaptation of the acoustic model, and unseen data for the evaluation of the adapted acoustic model:
\begin{equation}
    \mathcal{D} = \left \lbrace (D_{a_1}, D_{u_1}), \dots, (D_{a_N}, D_{u_N}) \right \rbrace.
\end{equation}
In theory, we should use an unlimited amount of unseen data for the evaluation of the adaptation algorithm. However, this is not practical when training the meta-learner. We approximate it by using $n$ seconds of speech as adaptation data and the following $n$ seconds of speech as unseen data. Further, as in any other machine learning problem we split the meta set $\mathcal{D}$ into a meta-training set $\mathcal{D}_{train}$ and a meta-validation set $\mathcal{D}_{val}$. The sets are split such that they contain different speakers, so that we can correctly assess the generalisation of the adaptation function to unseen speakers.

Finally, we use the loss $J$ to optimise the parameters $\Phi$ of the adaptation function using gradient descent:
\begin{equation}
    \hat{\Phi} = \arg\min_{\Phi} J \, .
    \label{eq:test_time_maml}
\end{equation}

\subsection{Speaker Adaptive Training as a Meta-Learning Task}

Speaker adaptive training is traditionally used to factor out speaker variation in order to enable a canonical model to focus on modelling phonological variances~\cite{anastasakos1996compact}. In model-based speaker adaptive training this may be done by splitting the weights of the acoustic model into a speaker-independent and a speaker-dependent set. During training, a copy of the speaker-dependent weights is maintained and optimised for each speaker separately~\cite{ochiai2015speaker,ochiai2016bottleneck,swietojanski2016sat}. Here, we take an alternative approach: Instead of maintaining and optimising a separate copy of speaker-dependent weights for each speaker we embed speaker adaptation directly into the acoustic model training using a meta-learning approach in order to find a good initialisation for speaker-dependent weights. The motivation for this is that the learned initialisation of the speaker-dependent weights ought to be more amenable to speaker adaptation compared to weights obtained through standard acoustic model training. Moreover, since this approach does not have to maintain a copy of speaker dependent weights for each speaker, it is possible to train all weights in a speaker adaptive fashion and allow the model to work out which weights should be adapted.

We described how to train a meta-learner for speaker adaptation above. To formulate speaker adaptive training as a meta-learning task, we jointly optimise the weights of the acoustic model $\Theta$ and the parameters of the meta-learner $\Phi$, minimizing the loss $J$ (Equation~\ref{loss}):
\begin{equation}
    \hat{\Theta}, \hat{\Phi} = \arg\min_{\Theta, \Phi} J
    \label{sat_loss}
\end{equation}

Speaker adaptive training using the meta-learning approach (SAT-MAML) is outlined in Algorithm~\ref{sat-maml}.

\subsection{Implementation of the adaptation function}

In order to train the adaptation function with Equation~(\ref{sat_loss}), the adaptation function must be differentiable. Previously~\cite{klejch2018learning}, the adaptation function was implemented as a coordinate-wise meta-learner~\cite{andrychowicz2016learning,ravi2016optimization} that uses a long short-term memory (LSTM) model~\cite{hochreiter1997long} to predict adapted weights for each weight $\theta \in \Theta$ of the acoustic model. More specifically, the LSTM predicts a forget gate $f$ and input gate $i$ that are used in an update rule for weight $\theta$ at time-step $t+1$:
\begin{equation}
    \theta_{t+1} = f \theta_{t} - i \nabla_{\theta_t} L(Y_a, f(X_a; \Theta_t))
\end{equation}

\input{algorithms/sat_maml.tex}
\input{algorithms/adapt.tex}

The LSTM uses three values as its inputs: the current weight value $\theta_t$, the current loss value $L(Y_a, f(X_a; \Theta_t))$ and the corresponding gradient $\nabla_{\theta_t} L(Y_a, f(X_a; \Theta_t))$. By using these three inputs the LSTM has expressive power to learn an adaptation schedule, and also to learn to escape local minima by resetting individual weights if the current loss is high and the corresponding gradients are small~\cite{ravi2016optimization}. In order to adapt all weights $\Theta$ of the acoustic model a large batch with dimension $|\Theta|\times3$ for each time-step is formed as an input to the LSTM. Unfortunately, training of the coordinate-wise meta-learner quickly becomes very memory inefficient for anything but a small number of speaker dependent weights because we have to store hidden states of the LSTM for all speaker dependent weights.

Given the computational complexity of the coordinate-wise meta-learner, we implemented the adaptation function as in model-agnostic meta-learning (MAML)~\cite{finn2017model}. This is more memory- and computationally-efficient, yet it has the same modelling power as the coordinate-wise meta-learner~\cite{finn2018meta}. With MAML the adaptation function is implemented as $n$ steps of gradient descent, where the set of learnable parameters $\Phi$ consists only of the learning rate $\alpha$, such that $\Phi = \left \lbrace \alpha \right \rbrace$. A single step of the adaptation function (Algorithm~\ref{adapt}) is then implemented as:
\begin{equation}
    \adapt(f, \Theta, D_a; \left \lbrace \alpha \right \rbrace) = \Theta - \alpha \nabla_{\Theta} L(Y_a, f(X_a; \Theta)).
\end{equation}

Note that in the original MAML implementation~\cite{finn2017model} the learning rate $\alpha$ was predefined and fixed during training, but it has been shown that the learning rate $\alpha$ can be trained jointly with the weights $\Theta$~\cite{antoniou2018train}. Also note that meta-learning approaches use second order derivatives to train the weights $\Theta$ and the parameters $\Phi$ of the adaptation function. It has, however, been demonstrated that using only the first order derivatives also works,  and this is faster and more stable during training~\cite{ravi2016optimization,finn2017model}. Hence, we chose to use only the first order derivatives in our experiments.

\subsection{Batch normalisation in MAML}
Since all our baseline models use batch normalisation~\cite{ioffe2015batch}, we also wanted to use it in the MAML models. In order to explain why using batch normalisation in MAML is complicated let us first briefly describe how batch normalisation works. Batch normalisation normalises hidden activations $h$ in the following way:
\begin{equation}
    h' = \gamma \frac{h - \mu}{\sqrt{\sigma^2 + \epsilon}} + \beta,
\end{equation}
where $\gamma$ and $\beta$ are the learned scale and offset weights, $\mu$ and $\sigma^2$ are the mean and variance statistics estimated on the current batch during training (denoted $\mu_B$ and $\sigma^2_B$) or as running statistics during inference (denoted $\mu_G$ and $\sigma^2_G$), and $\epsilon$ is a small number preventing division by 0. Previous papers used batch normalisation with mean and variance statistics computed only on the current batch~\cite{finn2017model} or accumulated different running statistics for each training step, because there was a big shift in distributions of hidden activations between different training steps~\cite{antoniou2018train}. 

We believe that using statistics computed only on the current batch is not optimal, because it forces the model to perform batch normalisation per utterance during inference. And since each utterance might have different duration, we would be using inconsistent estimates of true mean and variance for each utterance -- it might be compared to performing cepstral mean and variance normalisation per utterance instead of cepstral mean and variance per speaker. This also hurts performance of the speaker adaptation because we are adapting parameters with different statistics than statistics that will be used during inference. Therefore we decided to use batch renormalisation~\cite{ioffe2017batch}. Batch renormalisation normalises hidden activations $h$ in the following way:
\begin{equation}
    h' = \gamma \left ( r \frac{h - \mu_B}{\sqrt{\sigma_B^2 + \epsilon}} + d \right) + \beta,
    \label{eq:renorm}
\end{equation}
where $r$ and $d$ are computed during forward pass as follows:
\begin{equation}
    r = \frac{\sqrt{\sigma_B^2 + \epsilon}}{\sqrt{\sigma_G^2 + \epsilon}},
    \label{eq:r}
\end{equation}
\begin{equation}
    d = \frac{\mu_B - \mu_G}{\sqrt{\sigma_G^2 + \epsilon}}
    \label{eq:d}
\end{equation}
and are treated as constants during backward pass. This allowed us to accumulate running statistics $\mu_G$, $\sigma^2_G$ during training and adapt the model with respect to the same hidden activations during adaptation and inference, because by substituting Equations~\ref{eq:r}~and~\ref{eq:d} into Equation~\ref{eq:renorm} we obtain:
\begin{equation}
    h' = \gamma \frac{h - \mu_G}{\sqrt{\sigma_G^2 + \epsilon}} + \beta.
\end{equation}

\section{Experiments}


We performed experiments on TED talks because they provide many single speaker recordings, which allowed us to train models in a speaker adaptive way. The models were trained
using the TED-LIUM dataset~\cite{rousseau2012ted} and evaluated on TED talks that were used for the IWSLT~2010, 2011 and 2012 evaluations~\cite{iwslt2010, iwslt2011, iwslt2012}. To comply with the evaluation protocol, we used only TED talks that had been recorded before the end of 2012 for model training.  This resulted in 130 hours of raw training data.  We increased the size of the training data by a factor of 3 by performing speed perturbation~\cite{ko2015audio}. The development (dev) set consists of 18 speakers with an average speech duration of 6.5 minutes and the test set consists of 30 speakers with an average speech duration of 10.8 minutes. The training scripts were implemented in Tensorflow~\cite{abadi2016tensorflow} and Keras~\cite{chollet2015keras} and we used Kaldi~\cite{povey2011kaldi} for decoding. The scripts are publicly available.\footnote{\url{https://github.com/ondrejklejch/learning_to_adapt}}

\subsection{Speaker adaptation}

As a baseline we performed speaker adaptation of LHUC weights and ALL weights of a baseline model. In both methods we used 3 steps of full-batch gradient descent to adapt the weights. In all experiments we used the dev set to train a meta-learner to find the per-layer learning rate for 10s, 30s and 60s of adaptation data. We performed both supervised and unsupervised speaker adaptation experiments. Unsupervised labels were obtained using a separately trained baseline model and were used for unsupervised speaker adaptation of all models. During adaptation, we removed frames corresponding to silence from the adaptation data, because silence does not contain any speaker information.

\subsection{Baseline model}

All models used LDA projected~\cite{duda2012pattern} 40 dimensional MFCC features without cepstral mean and variance normalisation. The model was a time-delay neural network (TDNN)~\cite{peddinti2015time} and it consisted of 6 hidden layers with 600 neurons and a ReLU activation function~\cite{nair2010rectified} followed by batch normalisation~\cite{ioffe2015batch} and an LHUC layer~\cite{swietojanski2016learning}. LHUC layers were used only for adaptation. The model predicted posteriors of 4208 tied context dependent phones. The model had 5.9M parameters in total.\footnote{Note, that this model is 50\% smaller than a model used in a corresponding Kaldi recipe that has 850 neurons in each hidden layer. But this smaller model has only 0.5\% absolute higher WER than the larger model.} We trained the model for 400 iterations with Adam~\cite{kingma2014adam}. In each iteration we trained on 2000 batches that contained 256 chunks with 8 frames. This roughly corresponds to doing 3 epochs of training on all available training data. We performed early-stopping~\cite{morgan1990continuous,prechelt1998automatic} by monitoring loss on a validation set after each iteration. We used the best performing model for decoding.

\subsection{SAT-LHUC}

The SAT-LHUC model had the same architecture and used the same training schedule as the baseline model except that during training we used speaker specific LHUC parameters for each speaker in order to make the model able to remove speaker variance. In order to obtain a speaker-independent model, we used speaker independent LHUC weights instead of speaker-dependent LHUC weights with probability 0.5, as suggested in the original SAT-LHUC paper~\cite{swietojanski2016sat}.

\subsection{MAML}

We trained a model with the same architecture using the MAML approach. For training of the model we used 10s of adaptation data and the goal was to improve performance on the following 10s of unseen data using three full-batch adaptation steps. We trained different models for adaptation of LHUC parameters (MAML-LHUC) and adaptation of ALL parameters (MAML-ALL). We trained the model with a combination of a loss computed with the original unadapted model and a loss computed with the adapted model. In our early experiments we found that using only the loss computed with the adapted model lead to worse results. We initialised the model with a baseline model that was trained for 200 iterations and continued training it for another 200 iterations using the MAML objective. We used Adam~\cite{kingma2014adam} as an optimiser. In each iteration we trained on 1024 batches that contained data for 4 different speakers. This way all models used the same number of frames for training. At the end we fine-tuned per-layer learning rates for varying amounts of adaptation data on the dev set. 

\section{Results}

All our models use batch normalisation. In our first experiment we tested how to treat the mean $\mu$ and variance statistics $\sigma^2$ of the batch normalisation when adapting the LHUC weights of the baseline model. Our results are shown in Table~\ref{table:batchnorm}. Traditionally, batch statistics $\mu_B$, $\sigma^2_B$ are used during training (we call them batch stats), and running mean $\mu_G$ and variance $\sigma^2_G$ statistics are used for inference (we call them global stats). The results in Table~\ref{table:batchnorm} suggests that for speaker adaptation with limited amounts of adaptation data it is better to use global statistics than batch statistics. This is likely because the global statistics are better estimates of the true means and variances -- by adapting LHUC weights we are essentially correcting for errors in the variance estimation. (This was the only experiment where we used a learning rate of 0.7 for adaptation of LHUC weights, because it was found to work well in~\cite{klejch2018learning}.) 

\input{tables/batchnorm.tex}
\input{tables/supervised.tex}
\input{tables/unsupervised.tex}

In the second experiment we conducted supervised speaker adaptation of the baseline model, SAT-LHUC, and MAML models (Table~\ref{table:supervised}). In all experiments we used adaptation schedules learned with the meta-learning approach~(Equation~\ref{eq:test_time_maml}), instead of a hand-crafted adaptation schedule.

When adapting all parameters, MAML-ALL typically outperforms the baseline, particularly for 60s of adaptation data. This suggests that it has found an improved schedule using MAML, compared to the hand-crafted schedule for Baseline-ALL.

In the experiments adapting the LHUC weights of the baseline model, using the learned adaptation schedule achieves much better results (the first column in Table~\ref{table:supervised}) than adaptation using a handcrafted learning rate schedule (the first column in Table~\ref{table:batchnorm}). This is because the learned learning rates are $3\times$ -- $4\times$ larger than the handcrafted learning rate 0.7, which was found to work well in~\cite{klejch2018learning}. This result has a straightforward explanation: In this paper we perform full-batch adaptation, whereas in~\cite{klejch2018learning}, adaptation was performed with a batch size of 256 frames. Consequently, adapting using a full batch of 10\,s (1000~frames) is approximately $4\times$ larger than a batch size of 256. Nevertheless, this highlights the usefulness of using the meta-learning approach for estimating the adaptation schedule compared to an excessive hyper-parameter search, which requires to select appropriate step-sizes and bounds on the hyper-parameters. We argue that the learned learning rates for LHUC layers are several orders of magnitude larger than commonly used learning rates. There is therefore a considerable chance that the hyper-parameter bounds would not be set optimally to uncover the best solution. 
We observed similar trends when we performed unsupervised speaker adaptation experiments (Table~\ref{table:unsupervised}), but obtained much higher WERs than with supervised speaker adaptation.

When we compare the baseline model with SAT-LHUC and MAML-LHUC models we see that there is not a big difference between adaptation of these models. There are several possible explanations for this observation.
First, SAT-LHUC was originally shown to improve speaker adaptation with a feed-forward neural network~\cite{swietojanski2016sat}. This feed-forward neural network was approximately $5\times$ bigger than our TDNN baseline, but it achieved much worse performance before adaptation. It is possible that our TDNN baseline is sufficiently powerful to model both speaker and phonological variability inside the canonical model. It therefore might not benefit from factoring out speaker variability into speaker dependent weights as much as the feed-forward neural network. Second, the feed-forward neural network used in~\cite{swietojanski2016sat} did not employ batch normalisation. It is possible that the batch normalisation may implicitly be removing speaker variation from the data, thus removing the need for speaker adaptation. Third, there are only 881 speakers in the training data and differences between them are probably much smaller than differences between different classes in the few-shot learning scenario~\cite{finn2017model}, therefore the model is not forced to factor out speaker variability into speaker dependent weights. We plan to address all these hypotheses in our future work by running experiments with different model architectures, with and without batch normalisation, and by using data augmentation to introduce greater speaker variability into the training set.


\section{Conclusions}

In this paper we have extended the meta-learning approach for speaker adaptation~\cite{klejch2018learning} to support speaker adaptive training. The aim of speaker adaptive training is to remove speaker variability from the data in order for the canonical acoustic model to focus solely on modelling phonetic variation. Previous speaker adaptive training approaches used speaker-dependent weights to factor out speaker variability during training.  Rather than removing speaker variation during training, we proposed to use meta-learning to embed speaker adaptation via gradient descent directly into the training of the acoustic model. We hypothesised that this should result in the acoustic model learning weights that are more amenable to test-time adaptation, because the optimisation process using the meta-learning objective should steer weights of the acoustic model to regions that allow rapid adaptation. However, in our experiments we found that speaker adaptive training -- using SAT-LHUC or MAML -- did not improve performance of speaker adaptation of a strong state-of-the-art baseline model.

We showed that using global statistics for batch normalisation is beneficial compared to using statistics computed on the adaptation data in the low data adaptation regime, because global statistics are better estimates of the true mean and variance. We then demonstrated that using the meta-learner for estimation of an adaptation schedule achieves good results without any need for an excessive hyper-parameter search that still requires good bounds on the hyper-parameters. 

In future work we plan to further analyse why speaker adaptive training of current state-of-the-art models does not yield improvements. We would also like to devise a method that will perform better in rapid adaptation and unsupervised adaptation. Finally, we plan to evaluate meta-learning on larger tasks such as domain adaptation or language adaptation of ASR models.

%% file: algorithms/sat_maml.tex
\begin{algorithm}[t]
\caption{SAT-MAML}\label{sat-maml}
\begin{algorithmic}[1]
\Require \text{Number of iterations}
\Require \text{Number of speakers per batch}
\Require \text{Number of adaptation steps}
\Require \text{Learning rate $\alpha$}
\Function{SAT-MAML($f, \mathcal{D})$}{}
\State $\Theta \gets \text{random initialization}$
\State $\Phi \gets 0.001$

\For{$i \in \left \lbrace 1 \cdots iterations \right \rbrace$}
    \For{$s \in \left \lbrace 1 \cdots speakers\_per\_batch \right \rbrace$}
        \State  $D_a, D_u \gets $ sample$(\mathcal{D})$
        \State $J_s \gets L(Y_u, f(X_u; $ ADAPT$(f, \Theta, D_a; \Phi)))$
    \EndFor
    \\
    \State $J \gets \sum J_s $
    \State $\Theta \gets \Theta - \alpha \nabla_\Theta J$
    \State $\Phi \gets \Phi - \alpha \nabla_\Phi J$
\EndFor
\\
\\
\Return $\Theta, \Phi$
\EndFunction
\end{algorithmic}
\end{algorithm}

%% file: algorithms/adapt.tex
\begin{algorithm}[t]
\caption{Adaptation function}\label{adapt}
\begin{algorithmic}[1]
\Function{$\text{adapt}(f, \Theta, D_a; \Phi)$}{}
    \State $\Theta_0 \gets \Theta$
    \For {$j \in \left \lbrace 1 \cdots adaptation\_steps \right \rbrace$}
        \State $L_j \gets L(Y_a, f(X_a; \Theta_{j-1}))$
        \State $\Theta_j \gets \Theta_{j-1} - \Phi \nabla_{\Theta_{j-1}} L_j$
    \EndFor\\ 
    \Return $\Theta_{adaptation\_steps}$
\EndFunction
\end{algorithmic}
\end{algorithm}

%% file: tables/batchnorm.tex
\begin{table}[t!]
\begin{center}
\setlength{\tabcolsep}{1.1em}
\begin{tabularx}{\columnwidth}{Xcccc}
                        & \multicolumn{2}{c}{{global stats}} & \multicolumn{2}{c}{{batch stats}} \\
						&  {dev} & {test} &  {dev} & {test}  \\
\hline
original        		& 15.3 & 13.4 & 15.3 & 13.4 \\
10s			            & 15.1 & 13.1 & 15.2 & 13.3 \\
30s    		            & 14.8 & 12.7 & 15.0 & 13.2 \\
60s			            & 14.5 & 12.3 & 14.8 & 13.0 \\

\end{tabularx}

\caption{WERs (\%) of supervised speaker adaptation of LHUC parameters of the baseline model. Batch normalisation statistics are estimated on all the training data (global stats), or on the adaptation data (batch stats). The global statistics are likely more robust given the small amounts of data, producing lower error rates.}

\label{table:batchnorm}
\end{center}
\end{table}

%% file: tables/supervised.tex
\begin{table*}[t!]
\begin{center}
\setlength{\tabcolsep}{1.1em}
\begin{tabularx}{\textwidth}{Xcccccccccc}
                        & \multicolumn{2}{c}{Baseline-LHUC} & \multicolumn{2}{c}{Baseline-ALL} & \multicolumn{2}{c}{SAT-LHUC} & \multicolumn{2}{c}{MAML-LHUC} & \multicolumn{2}{c}{MAML-ALL} \\
						&  dev & test &  dev & test &  dev & test &  dev & test &  dev & test \\
\hline
original				& 15.3 & 13.4 & 15.3 & 13.4 & 15.4 & 13.5 & 15.3 & 13.4 & 15.3 & 13.4 \\
10s 					& 14.8 & 12.7 & 14.7 & 12.6 & 14.9 & 12.6 & 14.8 & 12.7 & 14.7 & 12.5 \\
30s						& 14.5 & 12.2 & 14.2 & 12.0 & 14.4 & 12.1 & 14.4 & 12.2 & 14.3 & 11.9 \\
60s 					& 14.2 & 12.0 & 14.0 & 11.7 & 14.1 & 11.9 & 14.2 & 11.9 & 13.8 & 11.5 \\

\end{tabularx}

\caption{WER (\%) for supervised speaker adaptation of the baseline, SAT-LHUC, MAML-LHUC, and MAML-ALL models.}
\label{table:supervised}
\end{center}
\end{table*}

%% file: tables/unsupervised.tex
\begin{table*}[t!]
\begin{center}
\setlength{\tabcolsep}{1.1em}
\begin{tabularx}{\textwidth}{Xcccccccccc}
                        & \multicolumn{2}{c}{Baseline-LHUC} & \multicolumn{2}{c}{Baseline-ALL} & \multicolumn{2}{c}{SAT-LHUC} & \multicolumn{2}{c}{MAML-LHUC} & \multicolumn{2}{c}{MAML-ALL} \\
						&  dev & test &  dev & test &  dev & test &  dev & test &  dev & test \\
\hline
original				& 15.3 & 13.4 & 15.3 & 13.4 & 15.4 & 13.5 & 15.3 & 13.4 & 15.3 & 13.4 \\
10s 					& 15.1 & 13.0 & 15.1 & 13.0 & 15.2 & 12.9 & 15.1 & 13.0 & 15.0 & 13.0 \\
30s						& 15.0 & 12.9 & 14.9 & 12.9 & 15.1 & 12.8 & 15.0 & 12.8 & 15.0 & 12.9 \\
60s 					& 14.8 & 12.9 & 14.8 & 12.8 & 14.9 & 12.8 & 14.8 & 12.9 & 14.8 & 12.8 \\
\end{tabularx}

\caption{WER (\%) for unsupervised speaker adaptation of the baseline, SAT-LHUC, MAML-LHUC, and MAML-ALL models.}

\label{table:unsupervised}
\end{center}
\end{table*}